\documentclass{josis}
\usepackage{hyperref}
\usepackage[hyphenbreaks]{breakurl}
\usepackage{booktabs}
\usepackage{stmaryrd}
\usepackage[utf8]{inputenc}
\usepackage[T1]{fontenc}
\usepackage{cite}
\usepackage{alltt}
\usepackage{underscore}
\usepackage{changepage}

\usepackage{algorithm}
\usepackage{algpseudocode}

\usepackage{placeins}
\usepackage[table]{xcolor}
\definecolor{sienna}{RGB}{92, 31, 16}
\usepackage{amssymb,amsmath}
\usepackage{microtype}

\josisdetails{%
   number=20, year=2020, firstpage=xx, lastpage=yy, 
  doi={10.5311/JOSIS.YYYY.II.NNN},
   received={May 10, 2020}, 
   returned={Month DD, 2020},
   revised={Month DD, 2020},
   accepted={Month DD, 2020}, }

\makeatletter
\let\UrlSpecialsOld\UrlSpecials
\def\UrlSpecials{\UrlSpecialsOld\do\/{\Url@slash}\do\_{\Url@underscore}}%
\def\Url@slash{\@ifnextchar/{\kern-.11em\mathchar47\kern-.2em}%
    {\kern-.0em\mathchar47\kern-.08em\penalty\UrlBigBreakPenalty}}
\def\Url@underscore{\nfss@text{\leavevmode \kern.06em\vbox{\hrule\@width.3em}}}
\makeatother

\hypersetup{
colorlinks=true,
linkcolor=black,
citecolor=black,
urlcolor=black
} 

\runningauthor{\begin{minipage}{.9\textwidth}\centering Camara\end{minipage}}
\runningtitle{Big EO Data Semantics}

\begin{document}

\title{On the Semantics of Big Earth Observation Data for Land Classification }

\author{Gilberto Camara}\affil{National Institute for Space Research, Brazil}

\maketitle


\begin{abstract}
This paper discusses the challenges of using big Earth observation data for land classification. The approach taken is to consider pure data-driven methods to be insufficient to represent continuous change. We argue for sound theories when working with big data. After revising existing classification schemes such as FAO's Land Cover Classification System (LCCS), we conclude that LCCS and similar proposals cannot capture the complexity of landscape dynamics. We then investigate concepts that are being used for analyzing satellite image time series; we show these concepts to be instances of events. Therefore, for continuous monitoring of land change, event recognition needs to replace object identification as the prevailing paradigm. The paper concludes by showing how event semantics can improve data-driven methods to fulfil the potential of big data. 
\end{abstract}

\section{Introduction}

Satellite images are the most comprehensive source of data about our environment; they provide essential information on global challenges. Images provide information for measuring deforestation, crop production, food security, urban footprints, water scarcity, land degradation, among other uses. In recent years, space agencies have adopted open distribution policies. Petabytes of Earth observation data are now available. Experts now have access to repeated acquisitions over the same areas; the resulting time series improve our understanding of ecological patterns and processes\cite{Pasquarella2016}. Instead of selecting individual images from specific dates and comparing them, researchers can track change continuously \cite{Woodcock2020}. To handle big data, scientists are developing new algorithms for image time series (for recent surveys, see \cite{Gomez2016, Zhu2017, Zeng2020}). These methods are \emph{data-driven and theory-limited}. However, numbers do not speak for themselves\cite{boyd2012}. Data-driven approaches without solid theories can lead to results which will not increase our knowledge\cite{Kitchin2014}.

What could be the effect of data-rich research in Geography and GIScience? Miller and Goodchild\cite{Miller2015} state: 'data-driven research should support, not replace, decision making by intelligent and skeptical humans'. Kwan\cite{Kwan2016} recommends a critical evaluation of big data algorithms. Li et al.\cite{Li2016} call for fresh approaches to obtain 'causal and explanatory relationships from big spatial data'. In their view, theory-free or theory-poor models are not sufficient for conceptual advances in our knowledge of geographical reality. We need sound theories to deal with big data without drowning in it.

Consider how experts use Earth observation data. Their input are images with resolution ranging from 5 to 500 meters, produced by satellites such as Landsat, Sentinels-1/2/3, and CBERS-4. To extract information, experts use methods that assign a label to each pixel (e.g., 'grasslands'). Labels can represent either \emph{land cover} or \emph{land use}. Land cover is the observed biophysical cover of the Earth's surface; land use concepts describe socio-economic activities\cite{Comber2008a}. Thus, `forest' is a type of land cover, while `corn plantation' is a kind of land use. To support land classification, scientists have proposed ontologies and descriptive schemes \cite{Herold2009}. We might thus ask: \emph{Are the current classification systems suitable to represent land change when working with big data? If not, which concepts are needed and how should they be applied?}

In what follows, we present the prevailing consensus on classification systems: FAO's Land Cover Classification System (LCCS)\cite{Herold2006a}. We argue that LCCS does not meet the challenges posed by big data. To support our views, we consider concepts used on image time series analysis; we show these concepys are related to event recognition and are not representable in LCCS. To improve the theory behind big data, we introduce elements of an event-centered ontology for land classification.

\section{Classification systems for Earth observation data: current status}

The act of classification raises philosophical questions dating as far back as Aristotle. We use an \emph{a priori} conception of reality to classify the world; what we observe has to fit our categories. Words in our language describe elements of the external reality. However, geographical terms such as `mountain' and `river' are imprecise and context-dependent\cite{Smith2001, Smith2003, Mark2003}. These ambiguities have motivated research on geospatial ontologies and semantics\cite{Smith1998, Fonseca2002, Kuhn2005}. In the early 2000s, there was much optimism about the impact of unified ontologies on modeling and interoperability of geographical information. However, building such complete ontologies is hard. Janowicz et al. \cite{Janowicz2012} recognize that "geographical concepts are situated and context-dependent, can be described from different, equally valid, points of view, and ontological commitments are arbitrary to a large extent".  Work on classification systems has shifted. Rather than using a single ontology, the current consensus argues for domain ontologies based on a common foundational ontology. These domain ontologies are means of making concepts of specific disciplines explicit and better communicating them\cite{Smith2007, Buttigieg2013}.

The semantics of Earth observation data are constrained by classification systems. Experts agree on what are the possible descriptions of the objects in the image (e.g., `forest', `river', `pasture').  Each pixel of the image is then labeled using visual or automated interpretation.  As an example, for countries reporting greenhouse gas inventories, the International Panel of Climate Change (IPCC) restricts the top-level land classes to `forest', `cropland', `grassland', `wetlands', `settlements', and `others'. This approach is too simplistic. Sasaki and Putz \cite{Sasaki2009} criticize the IPCC base classes for inducing wrong assessments for ecological and biodiversity conservation. The IPCC classes are an example where pre-conceived rules collide with the diversity of the world's ecosystems.

Since land classification provides essential information about our environment, many GIScience researchers have addressed the subject of land use and land cover semantics\cite{Comber2005, Ahlqvist2005, Ahlqvist2017}. They investigated consistency of classification systems \cite{Jansen2008}, semantic similarity between terms used by different systems\cite{Feng2004}, and disagreements between results\cite{Fritz2011}. The current consensus favors ontologies aiming at unambiguous definitions of land cover classes, such as the FAO Land Cover Classification System (LCCS)\cite{Herold2006a}. For this reason, it is important to discuss whether LCCS works well with big EO data.

FAO has developed the Land Cover Classification System (LCCS) "to provide a consistent framework for the classification and mapping of land cover"\cite{DiGregorio2016}. LCCS is a hierarchical system. At its highest level, LCCS has eight major land cover types:
\begin{enumerate}
	\item Cultivated and managed terrestrial areas.
	\item Natural and semi-natural terrestrial vegetation.
	\item Cultivated aquatic or regularly flooded areas.
	\item Natural and semi-natural aquatic or regularly flooded vegetation.
	\item Artificial surfaces and associated areas.
	\item Bare areas.
	\item Artificial water bodies, snow, and ice.
	\item Natural water bodies, snow, and ice.	
\end{enumerate}

The division on eight classes considers three criteria: presence of vegetation, edaphic conditions, and artificiality of cover\cite{DiGregorio2016}. Specialization of top-level LCCS classes uses properties such as life form, tree height, and vegetation density, setting pre-defined limits (e.g., "tree height > 10 meters"). These subdivisions are \emph{ad hoc} and application-dependent, leading to a combinational explosion with dozens or even hundreds of subclasses\cite{Herold2006a}. Such high expressive power can lead to incompatible LCCS-based class hierarchies\cite{Jansen2008}. 

LCCS is a landmark initiative; it provides a basis for a common understanding of land cover concepts. Many global and regional land mapping products use LCCS, including GLOBCOVER\cite{Arino2007} and ESA CCI Land Cover\cite{Li2016a}.  However, LCCS makes assumptions which limit its use with big data:

\begin{enumerate}
    \item LCCS describes land properties based only on land cover types, disregarding land use. For example, LCCS does not distinguish `pasture' from `natural grasslands'; it labels both as herbaceous land cover types. 
    \item The LCCS hierarchy uses hard boundaries between its subclasses. At each level of the hierarchy, properties of subclasses use fixed values (e.g., "sparse forests have between 10\% and 30\% of trees"). Real-world class boundaries do not fit into such strict definitions. When doing data analysis with machine learning, boundaries between classes are data-dependent and cannot be set a priori \cite{Hastie2009}. 
    \item Classification in LCCS has no temporal reference. LCCS assumes that subtype properties (e.g., percent of tree cover) are detectable at the moment of classification.  These properties do not refer to past or future values. Land use and land cover types whose values require time references (e.g., "forest land cleared in the last decade") are not representable in LCCS.  
\end{enumerate}

For example, the UNFCCC Reduction of Emissions by Deforestation and Degradation initiative (REDD+) requires representing and measuring forest dynamics\cite{Corbera2011}. Static and rigid definitions of `forest' used by LCCS cannot represent concepts such as `forest degradation'\cite{Putz2010}. Forest degradation happens when a natural forest loses part of its biodiversity and its tree cover. It is not a stable state but an intermediary situation that can lead to different medium-term outcomes. One can restore a degraded forest; degradation may continue and lead to complete loss of forest cover. Whatever the case, LCCS lacks explicit temporal information to capture forest degradation and thus support initiatives such as REDD+. Therefore, LCCS is thus not fit for many critical applications of EO data.

\section{Elements of an ontology of land use change}

\subsection{Overview} 

To represent change in geographical space, GIScience authors distinguish between \emph{continuants} and \emph{occurrents} \cite{Grenon2004, Galton2004, Galton2008, Worboys2005}. Continuants refer to entities that "endure through time even while undergoing different sorts of changes"\cite{Grenon2004}. The Amazon Forest and the city of Brasilia are continuants. Occurrents happen in a well-defined period and may have different stages during this time. Cutting down a forest area, cultivating a crop in a season, and building a road are occurrents. Objects are associated to continuants and events to occurrents.Philosophers have debated whether these two kinds of entities can be reduced to a single ontology or whether they are different perspectives of the same reality\cite{Simons2000}. We follow Grenon and Smith\cite{Grenon2004} who distinguish two top-level ontologies: one for continuants and another for occurrents. 
 
Atemporal classification systems such as LCCS refer only to properties of continuants. One can state facts such as "this area has 30\% forest cover" using LCCS, but cannot assert that "this area lost 70\% of its forest in the last two years". To convey change, classification systems for big data need to include occurrents. In what follows, we discuss concepts used in the analysis of satellite image time series. These time series are extracted from organized collections of Earth observation data covering a geographical area in regular temporal intervals. These concepts include `land-use change trajectory', `trend', `break', `disturbance', and `degradation'; they refer to occurrents and should be part of classification systems used in big EO data analysis. 
 
A caveat is in order. Philosophy of Language works such as Vendler\cite{Vendler1957} and Mourelatos\cite{Mourelatos1978} associate events to verbs (e.g., `run', `walk', `swim'). When dealing with land change analysis, we use nouns (e.g., `trend', 'break'). These nouns describe how properties measured in images change; as such, they represent occurrences. Although the proposed terms combine events with their measured properties, this is not a problem since the context is clear.  
 
\subsection{Land-use change trajectories}  
 
\begin{figure}[th]
\centering
\includegraphics[width=\textwidth]{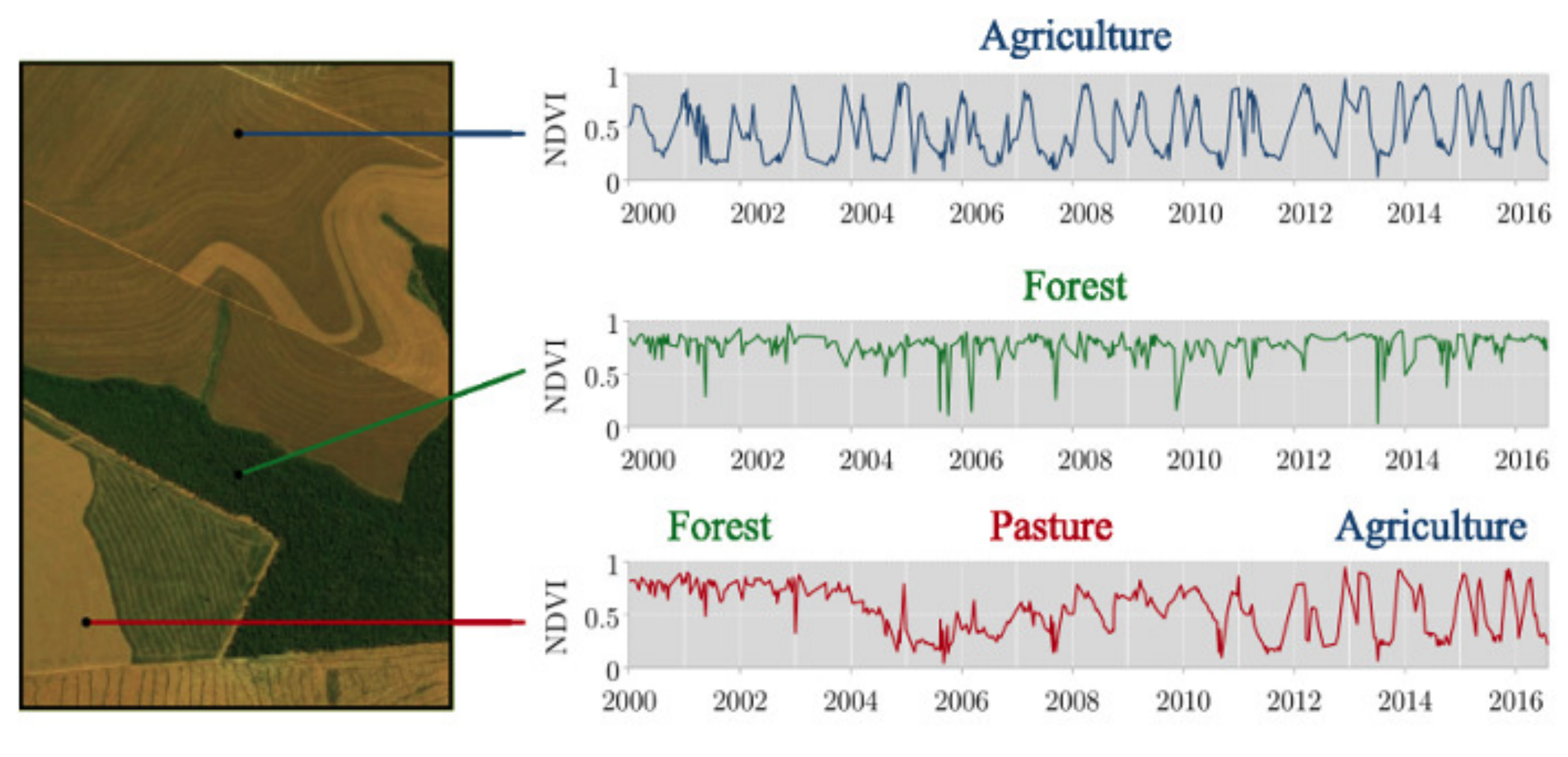}
\caption{Evolution of vegetation index in three locations in an agricultural region in Brazil, measured by MODIS sensor every 16 days from 2000 to 2017 (source: courtesy of Rolf Sim\~{o}es).}\label{fig:ts}
\end{figure}

As a foundational concept, we propose `land-use change trajectory' or 'trajectory' for short. A trajectory is a time series $l = \{(v_1, t_1), ..., (v_n, t_n)\}, v_i \in V, t_i \in T$. Each value $v_i$ from a domain $V$ is associated to an interval $t_i$ from a set $T$. Trajectories can be generic measures ("this dataset captures changes in one area from 2000 to 2020") or denote specific cases ("this dataset corresponds to an undisturbed forest from 2000 to 2010"). 

Consider Figure \ref{fig:ts}, which shows three locations in Brazil. On the left, a high-resolution image shows a snapshot of the area. The graphics show the NDVI vegetation index, measured by the MODIS sensor every 16 days from 2000 to 2017.  Spikes in the graphs are noise from clouds and can be ignored in the discussion. In the top graph, the vegetation index has a quasi-periodic variation, consistent with measures of agricultural areas. The middle graph shows a quasi-constant signal typical of a forest area. The bottom graphic reveals a more complex pattern. From 2000 to 2002, the signal is compatible with a forest. The vegetation index decreases from 2003 to 2005, indicating a deforestation event. From 2007 to 2010, the signal shows a small annual variation, which suggests the area was used as pasture for cattle raising. From 2011 to 2017, the index becomes similar to the top graph, which shows conversion to agriculture. In all cases, one can speak of land-use change trajectories measured by the sequence of satellite images.

\subsection{Patterns}

It is useful to identify `patterns', defined as trajectories $l_p$ representing a known event. Patterns $l_p$ are estimated from the data using statistical approximations\cite{Maus2016}, curve fitting\cite{Jonsson2002} or other adjustment methods\cite{Zeng2020}. Phenological models are important subclasses of patterns; they are useful for agricultural monitoring since they capture the onset, growth, maximum, and senescence of croplands\cite{Zeng2020}. For event recognition, researchers combine patterns with matching functions. Given a pattern $l_p$ and a trajectory $l$, a matching function $f_m(l_p, l)$ measures how much $l$ is similar to $l_p$. The most common matching functions are distance metrics $D(l_p, l)$ such as Euclidean distance and dynamic time warping\cite{Maus2016, Petitjean2012}. Pattern matching is the basis for many techniques for change detection using satellite images\cite{Zhu2017}.  

\begin{figure}[h]
\includegraphics[width=\textwidth]{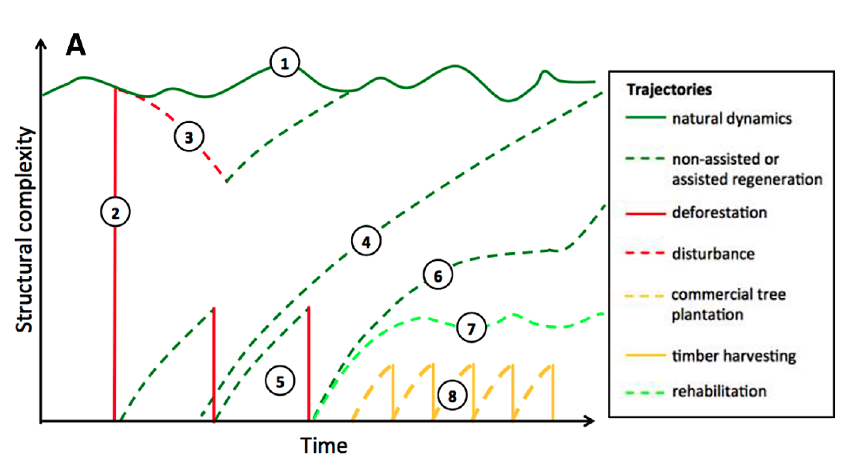}
\caption{Forest trajectories in terms of their structural complexity over time (source: Chazdon et al\cite{Chazdon2016}).}\label{fig:forest}
\end{figure}

Chazdon el al\cite{Chazdon2016} use patterns to address the question: \emph{when is a forest a forest?} The authors argue that current forest definitions used by institutions such as FAO and UNFCCC are incomplete. They state: "To assess and monitor forest and reforests properly requires viewing them as dynamic systems". The authors propose forest definitions based on patterns, as Figure \ref{fig:forest} shows. Different situations can occur, depending on how the structural complexity of forests varies in time. Trajectory (2) in Figure \ref{fig:forest} shows a complete loss of complexity, resulting in deforestation; trajectory (3) shows a case of forest degradation followed by a recovery. Other trajectories include the case of successful regeneration (case 4) and regeneration interrupted by a deforestation event (case 5). In all cases, the trajectories describe events. Therefore, understanding forest evolution requires event recognition.

Chazdon el at.\cite{Chazdon2016} point out that all trajectories shown in Figure \ref{fig:forest} are compatible with FAO's definition of forest: "Land with tree crown cover of more than 10\% and area of more than 0.5 ha. The trees should be able to reach a minimum height of 5 m at maturity in situ. Also includes areas normally forming part of the forest area which are temporarily unstocked as a result of human intervention or natural causes but which are expected to revert to forest". The FAO forest definition is an awkward attempt to take forest dynamics into account, resulting in confusion and leading to subjective interpretations. To support critical applications such as REDD+, FAO needs to update its definitions of 'forest' to include the temporal element.

\subsection{Labeled trajectories}

A labeled pattern is a trajectory $l = \{(v_1, t_1), ..., (v_n, t_n)\}$ to which we attach a description. The most common label is a class name such as 'forest' or 'grassland. Consider a trajectory described as "variation of vegetation index in location (-57.32, -11.56) from 2000 to 2010". By assigning a class label to the trajectory, we provide additional information about it. For example, if we find the area has been a forest area (see Figure \ref{fig:ts}) during this period, the trajectory's description becomes "location (-57.32, -11.56) was a forest from 2000 to 2010". 

As described in the previous section, patterns are useful for event recognition and thus suited for labeling trajectories. An alternative technique is machine learning, done in four phases: (a)  define the classification scheme; (b) select the time interval; (c) train the model; (d) classify the data. Users first define a set  of classes $C = \{c_1, \dots, c_n\}$. Then they choose a temporal interval for classification of time series (e.g., one year). The training data are sets of trajectories $T = \{L_1, \dots, L_n\}$; each trajectory fits into the chosen time interval and has a unique label. The result of the training is a function $f: L \rightarrow C$, which assigns a class $c$ to each trajectory $l$. Given the right conditions, machine learning produces good results for event recognition \cite{Pelletier2016, Picoli2018, Adarme2020}. 

Machine learning has limitations for event recognition. The machine learning function $f: L \rightarrow C$ assigns a unique class label $c$ to a trajectory $l$. This function only works for event recognition if each trajectory is homogeneous inside the chosen time interval. Consider the situation of the time series in the lower part of Figure \ref{fig:ts}. There is a transitional period from 2004 to 2006 when the forest was being removed and replaced by pasture. If training data does not include this transition, classification by machine learning will not recognize this event. Since most machine learning algorithms use fixed time intervals and hard boundary classes, they have limited explanatory power when dealing with transitional periods.

\subsection{Trends and breaks} 

Additional concepts for occurrents in time series of EO data include `trends' and `breaks'. A trend refers to long-term, average tendency of signals associated to an area on the ground\cite{Kennedy2010}. in most cases, trends are estimated using linear regressions\cite{Kennedy2010, Verbesselt2010}. Given a trajectory $l = \{(v_1, t_1), ..., (v_n, t_n)\}$, a trend is a linear approximation $l_v = \alpha + \beta*t_i, t_i \in \{t_1, \dots, t_n\}$. Trend analysis methods identify homogeneous subsets of a time series, supporting event recognition.

\begin{figure}[h]
\centering
\includegraphics[width=0.9\textwidth]{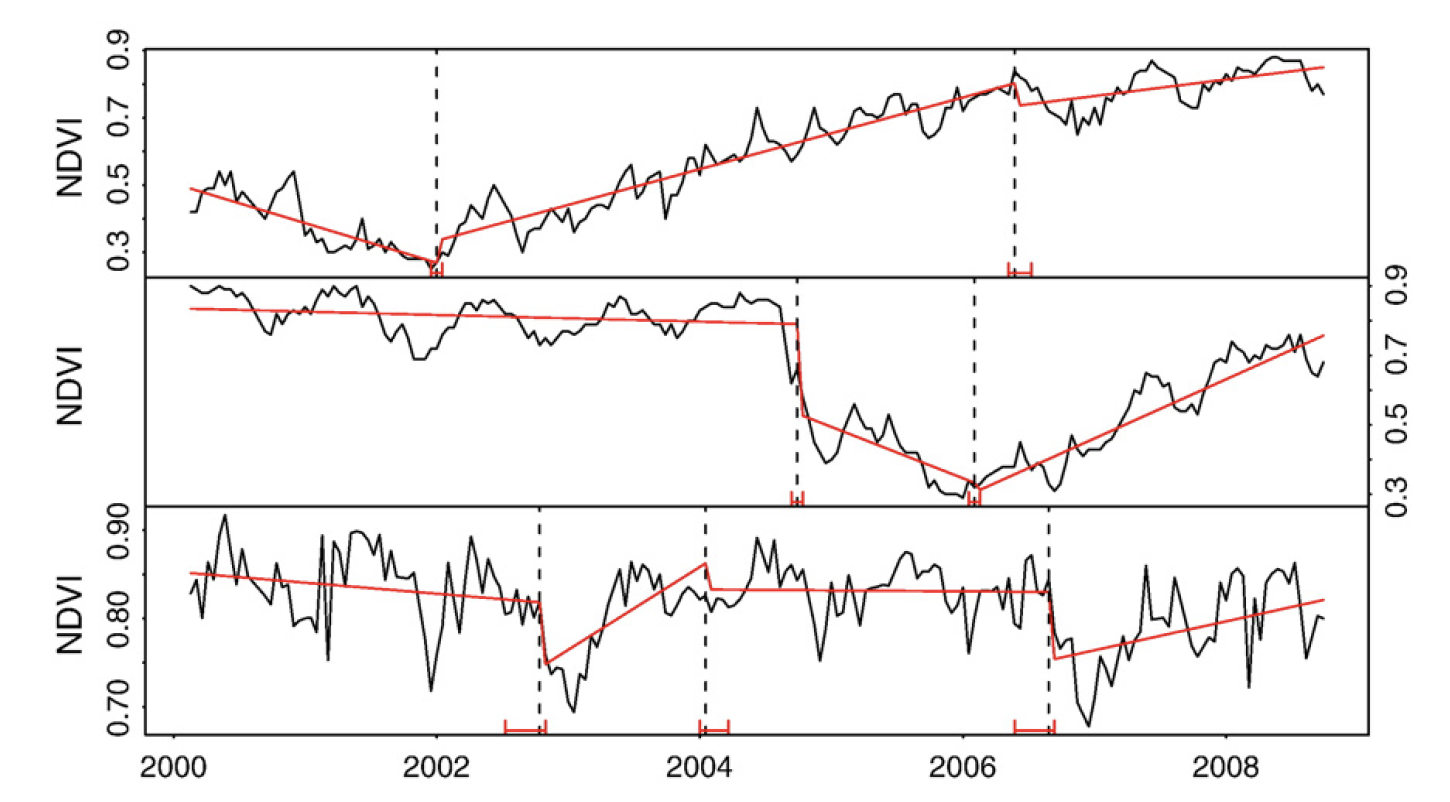}
\caption{Detected changes in the trend component (red) of 16-day NDVI time series (black) extracted from a single MODIS pixel within a pine plantation, planted in 2001 (top), harvested in 2004 (middle), and with tree mortality occurring in 2007 (bottom) (source: Verbesselt el al.\cite{Verbesselt2010}).}\label{fig:bfast}
\end{figure}

 A break signals abrupt changes in the time series and signals that a significant disturbance happened\cite{Verbesselt2010}. The OECD Glossary of Statistical Terms \cite{OECD2003},\ states: "breaks in statistical time series occur when there is a change in the standards for defining and observing a variable over time". Thus, breaks are instantaneous events associated to a discontinuity. Given a trajectory $l = \{(v_1, t_1), ..., (v_n, t_n)\}$, a break corresponds to an event $b$ associated to a value and a time $(v_b, t_b)$ that splits this trajectory in three parts:  $l = \{l_{p1}, b, l_{p2}\}$ where $l_{p1}$ and $l_{p1}$ are instances of two patterns, one occurring before and other after the break. Break detection is useful for event recognition; a break is a temporal boundary between two trajectories. Figure \ref{fig:bfast} shows an example of a break detection algorithm BFAST\cite{Verbesselt2010} applied to a MODIS time series of one pixel in a pine plantation. The granularity of the MODIS time series is 16 days. BFAST detects various breaks in the time series, associated to harvesting actions.

\subsection{A hierarchy of events for land classification}

\begin{figure}[h]
\centering
\includegraphics[width=0.7\textwidth]{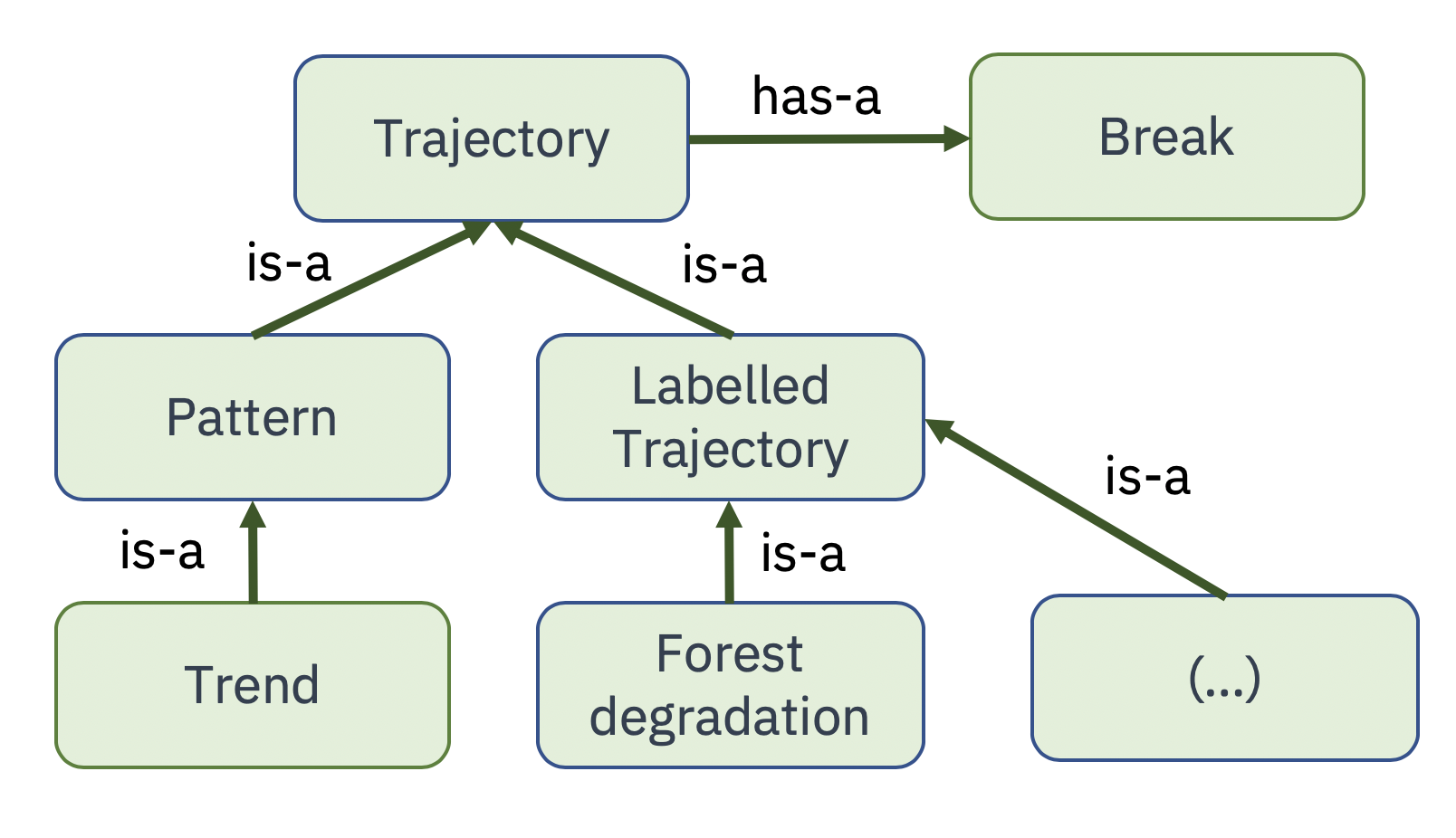}
\caption{Proposed hierarchy of concepts related to event recognition in satellite image time series (source: author).}\label{fig:events}
\end{figure}

In the previous sections, we examined concepts used for representing information about occurrents in satellite image time series. These concepts make up a hierarchy of types (see Figure \ref{fig:events}). Taking land-use change trajectory as the top concept, we propose `patterns` and `labeled trajectories' as subtypes of trajectories.  Breaks are different kinds of events. While trajectories span an interval, breaks are instantaneous considering the temporal granularity used in time series. More complex concepts such as `forest degradation' and `deforestation' are subtypes of 'labeled trajectories' which can be identified by pattern matching or by classification. For example, consider trajectory (3) in Figure \ref{fig:forest}, which Chazdon et al\cite{Chazdon2016} use as an example of forest disturbance. A pattern-matching algorithm would compare different time series with a pattern of forest disturbance. The alternative is to use machine learning with good training data. If the data has spatial and temporal resolution to capture the trajectories, both approaches are adequate for event recognition. Experts can extend the hierarchy further by identifying additional types of events. 

\section{Conclusion: event recognition as a basis for big Earth observation data analysis}

In this paper, we discuss the challenge of supporting big data analysis with sound theory. We argue that time series analysis, including pattern matching, trend analysis, break detection, and time series classification, are subtypes of \emph{event recognition}. When doing continuous monitoring of land change, it is not advisable to use LCCS and similar approaches. Instead of identifying classes such as `forest' or `grasslands', data analysis methods need to recognize events such as "this area was a forest from 2000 to 2010, then it was deforested in 2011, and turned into grasslands from 2011 to 2020". When doing continuous monitoring, event recognition replaces object identification as the purpose of land classification. 

The emphasis on event recognition has significant consequences for the design of algorithms and classification systems for big data.  In particular, machine learning is not a panacea. Continuous monitoring of land dynamics using remote sensing data differs from applications such as spam filters, automatic translation, and object detection. Land systems do not change overnight. There is a period of transition for land cover conversion. Depletion of natural resources such as forests and wetlands can take place over months or even years. Monitoring subtle land transitions is crucial for protecting our environment. Thus, machine learning methods need to be adapted to work with satellite image time series.

Most algorithms for big EO data analysis use techniques that have proven useful in other problems. However, monitoring natural resources is more complex than detecting spam emails or playing chess. While machine learning and pattern analysis are useful, there is still much to be done to build sound theories for dealing with big EO data. Long-term progress will depend on a new generation of methods that combine machine learning with functional ecosystem models. A better theoretical basis is essential for algorithms that extract information from petabytes of free EO data. This new generation of combined methods will allow a better understanding of the processes that drive landscape dynamics.

\section*{Acknowledgments}

The author would like to thank his peers at Brazil's National Institute for Space Research (INPE) for many useful discussions, including Isabel Escada, Karine Ferreira. Lubia Vinhas, Michelle Picoli, Alber Sanchez, Claudio Almeida, Gilberto Queiroz, Miguel Monteiro, and Rolf Simoes.

\bibliographystyle{josisacm}

\end{document}